\documentclass[]{beingbeyond}

\usepackage[toc,page,header]{appendix}

\title{RealDexUMI: A Wearable Universal Manipulation Interface 
for Dexterous Robot Learning}

\author{{\bfseries
Chaoyi Xu$^{1,2}$~
Yixuan Jiang$^{3}$~
Jiahui Huan$^{2}$~
Yuhui Fu$^{1,2}$~
Haoyu Zhou$^{4}$ \\
Weitian Yuan$^{4}$~
Jiayi Yu$^{2,5}$~
Wanpeng Zhang$^{1,2}$~
Haoqi Yuan$^{1,2}$~
Zongqing Lu$^{1,2,\dagger}$
}}

\affiliation{
$^{1}$Peking University \quad
$^{2}$BeingBeyond \quad
$^{3}$Beihang University \quad
$^{4}$LinkerBot \quad
$^{5}$Tsinghua University \quad
}

\webpage{\url{https://research.beingbeyond.com/realdexumi}}

\usepackage{amsmath,amssymb}
\usepackage{enumitem}
\usepackage{array}
\usepackage{pifont}
\usepackage{adjustbox}
\usepackage{float}

\newcommand{\acknowledgments}[1]{}

\newcommand{\cmark}{\textcolor{green!60!black}{\ding{51}}}
\newcommand{\xmark}{\textcolor{red!70!black}{\ding{55}}}

\newcolumntype{L}[1]{>{\raggedright\arraybackslash}m{#1}}
\newcolumntype{C}[1]{>{\centering\arraybackslash}m{#1}}

\abstract{
Learning dexterous manipulation requires demonstrations that preserve fine hand-object interactions while remaining executable at deployment. 
Existing pipelines either lose deployable dexterity through retargeting or embodiment conversion, 
or rely on robot-specific teleoperation that is costly to scale and often lacks intuitive, contact-aware control for dexterous data collection. 
We present RealDexUMI, a wearable universal manipulation interface built around a shared dexterous end-effector module that integrates a lightweight dexterous hand, in-hand vision, and fingertip tactile sensing.
A palm-side isomorphic teleoperation glove maps human finger inputs to robot-hand joint commands, enabling real-time, retargeting-free, intuitive, and precise hand control. 
The shared hand and sensing modules yield zero-gap end-effector data, with matched in-hand observations, tactile signals, contacts, and hand actions between collection and deployment.
Across eight real-robot tasks spanning fine-grained, contact-rich, long-horizon, and bimanual manipulation, policies trained on RealDexUMI data achieve an average success rate of 88.75\%, generalize to unseen initial poses, and transfer across three embodiments.
}

\checkdata[Date]{Jun 4, 2026}

\makeatletter
\g@addto@macro\@thanks{%
\footnotetext[2]{Correspondence to Zongqing Lu $<$lu@beingbeyond.com$>$.}%
}
\makeatother

\firstfig[
    width=1.0\linewidth,
    trim=0mm 40mm 0mm 0mm,
    clip
][\linewidth]{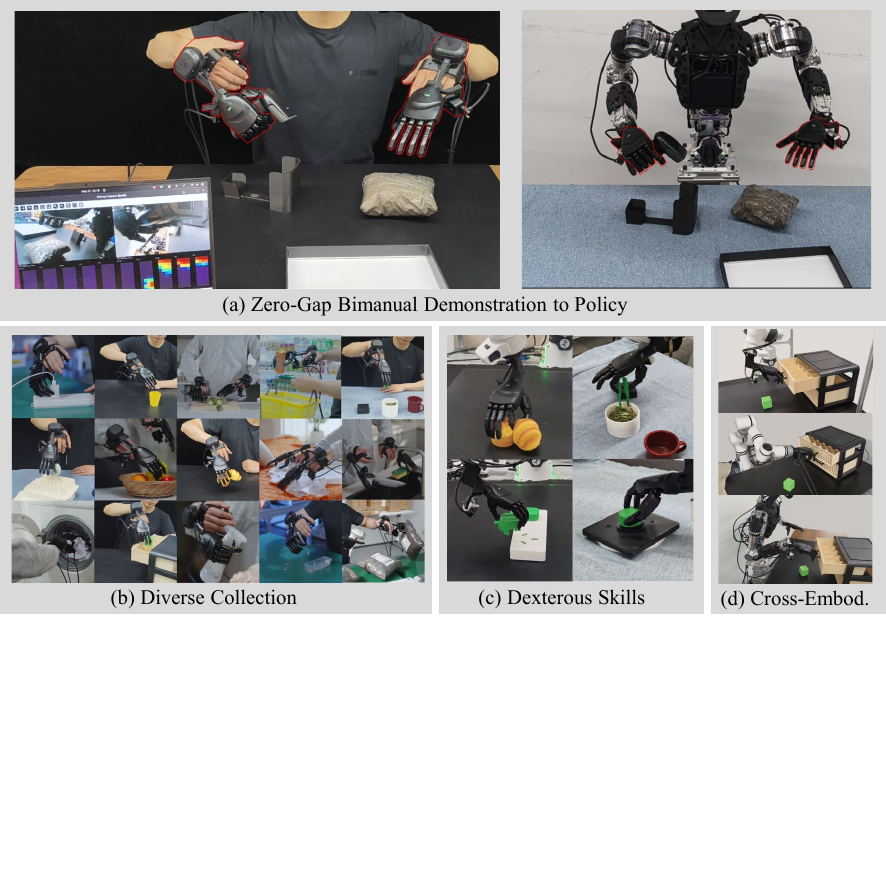}{
RealDexUMI turns a dexterous hand into a shared interface for zero-gap wearable demonstration and robot deployment (a), supporting diverse task collection (b), dexterous skill learning (c), and cross-embodiment deployment (d).
}{fig:teaser}

\begin{document}
\raggedbottom
\maketitle


\section{Introduction}

Human dexterity in demonstrations does not automatically translate into deployable robot dexterity. 
Many demonstration interfaces can capture dexterous human behavior, but the resulting data may still differ from the hand actions the robot can execute, the contacts it makes, and the observations it receives at deployment~\cite{chen2026dexvitac,xu2025dexumi,wu2025robocoin,yang2024ace,fang2025airexo,beingbeyond2025beingh0,beingbeyond2026beingh05}. 
For dexterous manipulation, the relevant quantity is therefore not captured dexterity alone, but \textbf{deployable dexterity}: the extent to which demonstrated hand actions remain executable by the deployed dexterous end effector while the associated contacts, tactile signals, and observations are preserved.
This distinction is crucial for contact-rich dexterous manipulation, where small deviations can determine whether a skill succeeds or fails.

Existing data-collection pipelines preserve deployable dexterity only partially. 
Hand-motion interfaces can capture detailed finger motions or use them to teleoperate robot hands, but both rely on human-to-robot mappings across different kinematics, contact geometries, and sensing channels~\cite{chen2026dexvitac,wu2025robocoin,tao2025dexwild,zheng2025world,cheng2024open,iyer2024open,qin2023anyteleop,unitachand}.
Such offline or online retargeting can distort contact-rich interactions and make precise robot-hand behaviors difficult to produce during collection.
Robot-specific leader-follower teleoperation reduces this mapping ambiguity, but ties data collection to robot-specific hardware and does not naturally scale to wearable, cross-embodiment dexterous data collection~\cite{fu2024mobile,wu2023gello,bytedance2025bytedexter}.
This motivates an interface that supports efficient collection while remaining faithful to the deployed dexterous end effector.

\textbf{RealDexUMI} follows a different principle: preserve deployable dexterity by using a shared dexterous end-effector module as both the wearable collection interface and the deployed robot hand.
The module integrates a lightweight dexterous hand, in-hand vision, and fingertip tactile sensing, while a palm-side isomorphic glove provides direct commands in the hand's executable action space.
A relative end-effector action representation further lets the same policy operate across embodiments by changing only the inverse kinematics (IK) and low-level controller.

Experiments across eight real-robot tasks, collection-time interface comparisons, ablations, and cross-embodiment deployment validate RealDexUMI as a scalable interface for deployable dexterous policy learning.

\vspace{2mm}

We make three contributions:
\begin{itemize}[leftmargin=*]
    \item \textbf{A wearable interface for deployable dexterity.}
    We introduce RealDexUMI, a wearable device built around a shared dexterous end-effector module and a palm-side isomorphic glove for real-time, retargeting-free, and precise control of the hand module, enabling reliable collection of dexterous demonstrations in the wild.

    \item \textbf{Zero-gap dexterous data from collection to deployment.}
    By using the same dexterous end-effector module for collection and execution, RealDexUMI preserves observations, contact surfaces, executable hand commands, and action--state correspondence.

    \item \textbf{Cross-embodiment dexterous policy deployment.}
    Policies trained on RealDexUMI data use relative end-effector actions to deploy the same checkpoints across embodiments without retraining and remain robust to unseen initial poses.
\end{itemize}


\begin{table}[t]
\centering
\small
\setlength{\tabcolsep}{2.7pt}
\renewcommand{\arraystretch}{1.05}
\caption{
Comparison of representative demonstration interfaces.
}
\label{tab:system_comparison}
\begin{adjustbox}{max width=1.0\linewidth}
\begin{tabular}{
L{2.3cm}
C{1.5cm}
C{1.5cm}
C{1.5cm}
C{1.5cm}
C{1.0cm}
C{1.0cm}
C{1.0cm}
C{1.0cm}
C{1.0cm}
}
\toprule
\textbf{System} &
\begin{tabular}[c]{@{}c@{}}
\textbf{Act.}\\[-2pt]
\textbf{DoF}
\end{tabular} &
\textbf{Weight} &
\begin{tabular}[c]{@{}c@{}}
\textbf{Teleop}\\[-2pt]
\textbf{complex.}
\end{tabular} &
\begin{tabular}[c]{@{}c@{}}
\textbf{Device}\\[-2pt]
\textbf{size}
\end{tabular} &
\begin{tabular}[c]{@{}c@{}}
\textbf{Free}\\[-2pt]
\textbf{wrist}
\end{tabular} &
\begin{tabular}[c]{@{}c@{}}
\textbf{Vision}\\[-2pt]
\textbf{align}
\end{tabular} &
\begin{tabular}[c]{@{}c@{}}
\textbf{No}\\[-2pt]
\textbf{retarg.}
\end{tabular} &
\textbf{Tac.} &
\begin{tabular}[c]{@{}c@{}}
\textbf{A--S}\\[-2pt]
\textbf{corr.}
\end{tabular} \\
\hline

UMI~\citep{chi2024universal} &
1 &
780 g &
Low &
Small &
\cmark &
\cmark &
\cmark &
\xmark &
\xmark \\

DexViTac~\citep{chen2026dexvitac} &
6 &
$<$600 g &
Low &
Small &
\cmark &
\xmark &
\xmark &
\cmark &
\xmark \\

DEXOP~\citep{fang2025dexop} &
12/9/7 &
$>$1000 g &
High &
Large &
\cmark &
\cmark &
\cmark &
\cmark &
\xmark \\

Dexexo~\citep{zhu2026dexexo} &
6 &
$>$1000 g &
High &
Large &
\xmark &
\cmark &
\cmark &
\xmark &
\xmark \\

DEX-Mouse~\citep{koh2026dex} &
6 &
$>$1000 g &
High &
Large &
\xmark &
\cmark &
\cmark &
\cmark &
\cmark \\

Exo-viha~\citep{chao2025exo} &
6 &
1100 g &
High &
Large &
\xmark &
\cmark &
\xmark &
\xmark &
\cmark \\

DexUMI~\citep{xu2025dexumi} &
12/6 &
850 g &
Med. &
Med. &
\cmark &
\xmark &
\cmark &
\cmark &
\xmark \\

\midrule
\textbf{RealDexUMI} &
\textbf{6} &
\textbf{680 g} &
\textbf{Low} &
\textbf{Med.} &
\textbf{\cmark} &
\textbf{\cmark} &
\textbf{\cmark} &
\textbf{\cmark} &
\textbf{\cmark} \\
\bottomrule
\end{tabular}
\end{adjustbox}
\par\vspace{4pt}
\begin{minipage}{\linewidth}
\footnotesize
\textit{Note.}
Act. DoF denotes actuated end-effector DoF; slash-separated values indicate hardware variants. 
Teleop complex. denotes perceived teleoperation complexity, obtained from a structured survey using reported system descriptions and publicly available demonstration materials. 
It reflects expected operator learning and control difficulty rather than hardware performance.
Vision align denotes collection--deployment visual alignment without nontrivial post-processing.
No retarg. denotes retargeting-free hand control.
Tac. denotes tactile sensing.
A--S corr. denotes action--state correspondence.
\end{minipage}

\end{table}

\section{Related Work}
\label{sec:relatedwork}

\subsection{UMI-Style Demonstration Interfaces}

UMI-style interfaces collect demonstrations without a robot body by using end-effector-centric observation and action representations~\citep{chi2024universal,seo2024legato}.  
This makes data collection scalable and robot-agnostic, and has inspired extensions with tactile sensing and additional visual viewpoints~\citep{lin2025data,liu2024fastumi,liu2025fastumi,huang20243d,zhu2025touch,xu2025exumi,cheng2026tacumi,luo2026omniumi,xu2026hommi,zeng2025activeumi,wang2026xrzero,liu2025vitamin,liu2024maniwav}. 
Related systems further extend this idea to mobile or whole-body manipulation~\citep{xu2026hommi,ha2024umilegs,yu2026bifrostumi,huang2026mobile}. 
However, most UMI-style systems use grippers whose limited actuation and simple contact geometry are insufficient for dexterous manipulation tasks.

\subsection{Dexterous Demonstration Interfaces}

Dexterous demonstration interfaces extend robot-free data collection from grippers to dexterous hands, with the key challenge of enabling intuitive control while preserving deployment-aligned observations, contacts, and hand commands.
Motion-capture gloves provide a natural way to capture rich human-hand motion~\citep{chen2026dexvitac,tao2025dexwild,zheng2025world,bunny-visionpro,wang2024dexcap,chen2024arcap,gao2025glovity}, but the recorded motion must be retargeted to robot hands with different kinematics and contact surfaces. 
Linkage-based exoskeletons obtain more robot-aligned joint states through mechanical coupling~\citep{fang2025dexop,zhu2026dexexo,wei2024wearable,zhang2025doglove,liang2026cdf,romero2024eyesight,jia2026feel,du2025mile,si2025exostartefficientlearningdexterous}, but measured hand states are not equivalent to executable hand commands for contact-rich interaction, and object contacts are not made through the deployed end effector.
Arm-attached systems~\citep{koh2026dex,chao2025exo} let operators carry a dexterous end effector directly, preserving robot-hand contacts and providing passive force feedback.
However, their large forearm-mounted structures restrict workspace and natural wrist motion, as reflected by the size and weight comparison in Table~\ref{tab:system_comparison}.
DexUMI uses a wearable hand exoskeleton to enable intuitive human-hand demonstrations with direct haptic feedback~\citep{xu2025dexumi}. 
However, its contacts are mediated by a human-worn exoskeleton rather than the deployed robot hand, the recorded signals are states rather than commands in the deployed hand’s action space
, and observation alignment still depends on a nontrivial visual-inpainting pipeline.

RealDexUMI follows the UMI principle with a compact wearable dexterous interface. 
Compared with prior dexterous interfaces, it preserves easy wearable control, deployment-aligned observations, contacts, tactile sensing, executable hand commands, and action--state correspondence without retargeting or visual post-processing.


\section{RealDexUMI}
\label{sec:realdexumi}

\begin{figure}[t]
    \centering
    \includegraphics[
        width=\linewidth,
        trim=5mm 25mm 5mm 25mm,
        clip
    ]{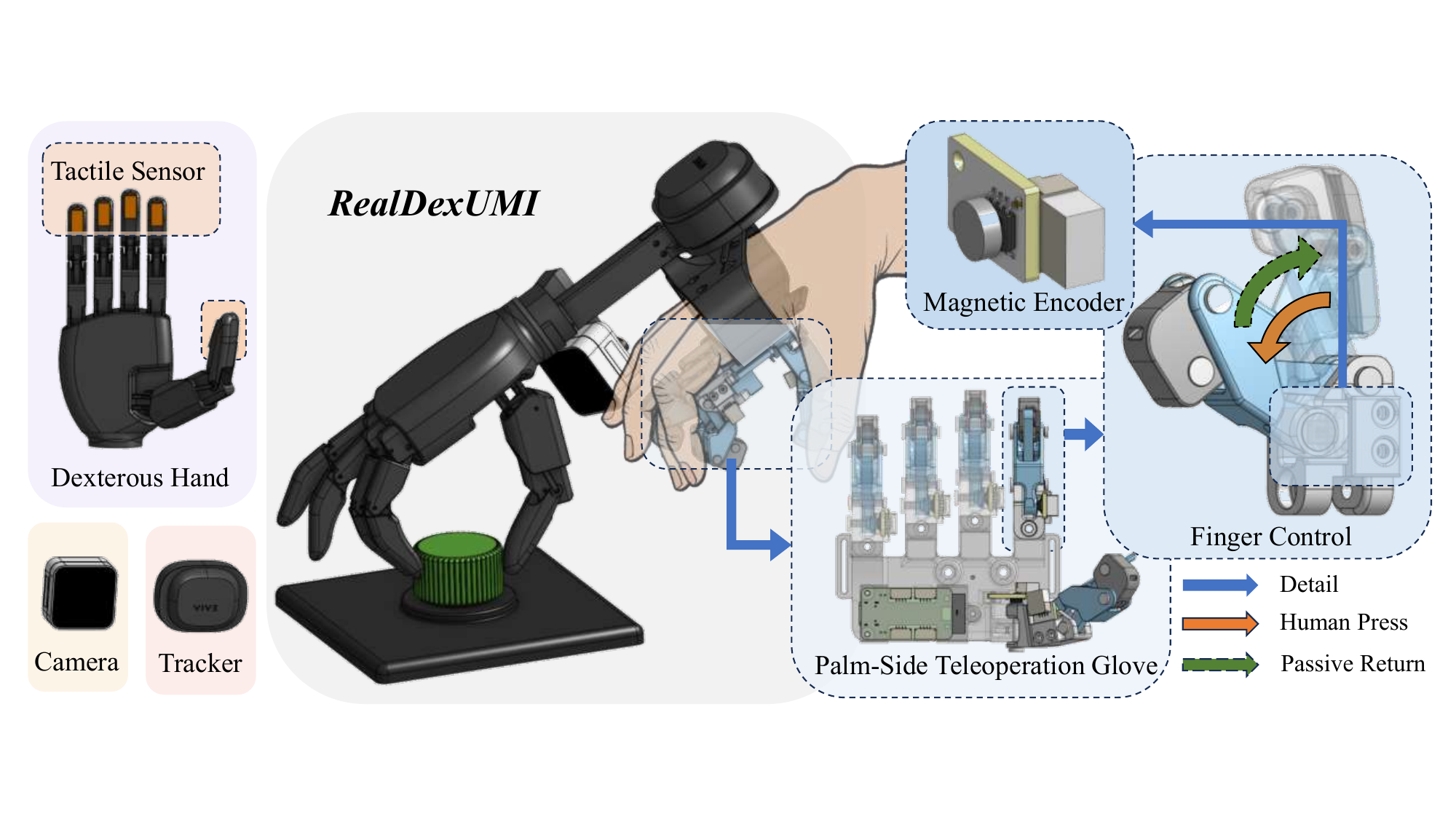}
\caption{
\textbf{Hardware system overview.}
The wearable device combines a reusable dexterous end-effector module, a 6-DoF tracker, and a palm-side isomorphic teleoperation glove.
The end-effector module consists of the lightweight dexterous hand, in-hand camera, and fingertip tactile sensors, and is mounted on robot bodies during deployment.
}
    \label{fig:hardware}
\end{figure}

\subsection{System Overview}

RealDexUMI is a wearable dexterous demonstration interface built around a reusable end-effector module. 
During collection, the operator wears the module and commands the deployable dexterous hand through a palm-side isomorphic glove, so object contact is made by the robot hand rather than the human hand.
During deployment, the same hand and in-hand camera are mounted as the robot end effector, allowing learned policies to operate with matched end-effector observations and executable hand commands.
For bimanual tasks, two synchronized RealDexUMI modules are used, and the policy predicts per-hand relative motions and hand commands from concatenated per-hand observations.
Together, these choices target wearable collection, retargeting-free hand control, and deployment-aligned end-effector data.

\subsection{Lightweight Dexterous Hand}

\begin{figure}[htbp]
  \centering
    \includegraphics[
        width=0.6\linewidth,
        trim=0mm 0mm 0mm 0mm,
        clip
    ]{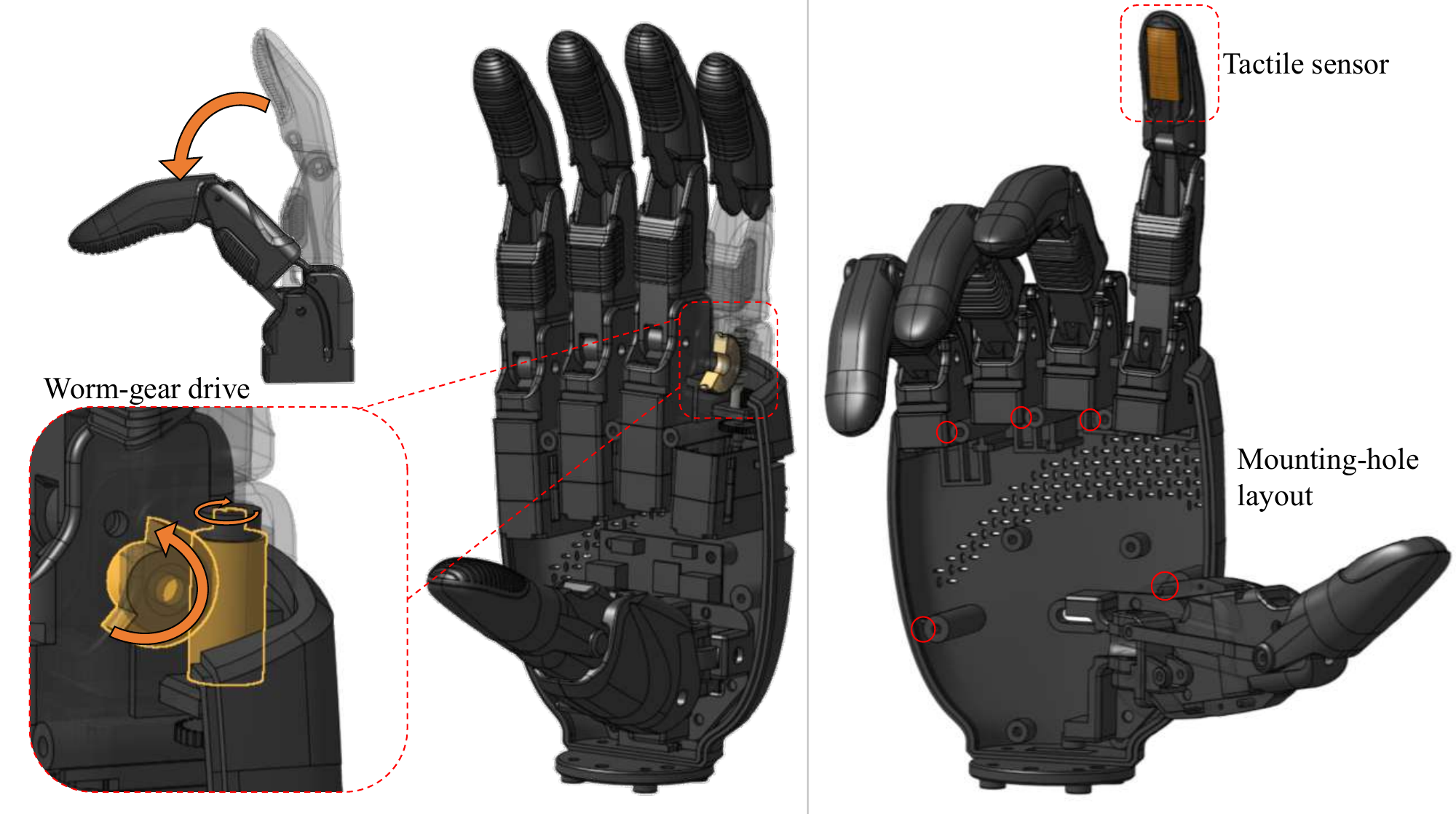}
\caption{
\textbf{Lightweight dexterous hand module.}
The hand uses compact finger actuation, integrated fingertip tactile sensing, and a lightweight structural shell.
}
  \label{fig:hand_design}
\end{figure}

As shown in Fig.~\ref{fig:hand_design}, the hand module is designed to make the end-effector interface wearable while retaining dexterous actuation and contact sensing.
The hand provides 11 degrees of freedom (DoFs), consisting of one actuated and one passive flexion DoF for each finger, as well as an additional actuated DoF for thumb abduction and adduction. 
Its six actuated DoFs use servo-driven worm-gear transmissions, allowing the hand to remain compact and lightweight.

To reduce structural mass, a machined high-toughness polycarbonate (PC) shell integrates actuator seats, finger mounts, and screw holes without separate support brackets.
Each fingertip integrates a piezoresistive tactile array.
These sensors provide explicit contact observations at the same fingertip surfaces used during robot execution, reducing reliance on vision-only contact inference.

\subsection{Palm-Side Isomorphic Teleoperation Glove}

The palm-side teleoperation glove is a robot-hand command interface rather than a human-hand motion-capture device.
Unlike mocap gloves that retarget human-hand motion to robot-hand commands,  RealDexUMI collects operator input directly in the deployed hand's executable command space.
As shown in Fig.~\ref{fig:hardware}, each of the six sensed glove DoFs corresponds to one actuated hand DoF, producing a 6-D command vector that directly matches the hand command space. 

The sensed DoFs are measured by absolute magnetic encoders and mapped linearly to hand commands, while five passive coupled DoFs mechanically mirror the hand's passive flexion DoFs.
The glove is fixed by a palm ring and operated by pressing mechanical links with the fingers.
This avoids full-hand exoskeleton wearing, keeps the operator's fingers unobstructed, and lets the robot hand itself make object contact.
Torsion springs and mechanical range limits return the glove to an open posture when released. 
Together with startup calibration, it provides real-time, precise, and retargeting-free control in the deployed hand command space.

\subsection{Demonstration Collection}

\begin{figure}[htbp]
  \centering
    \includegraphics[
        width=0.55\linewidth,
        trim=0.5mm 0mm 0mm 0.5mm,
        clip
    ]{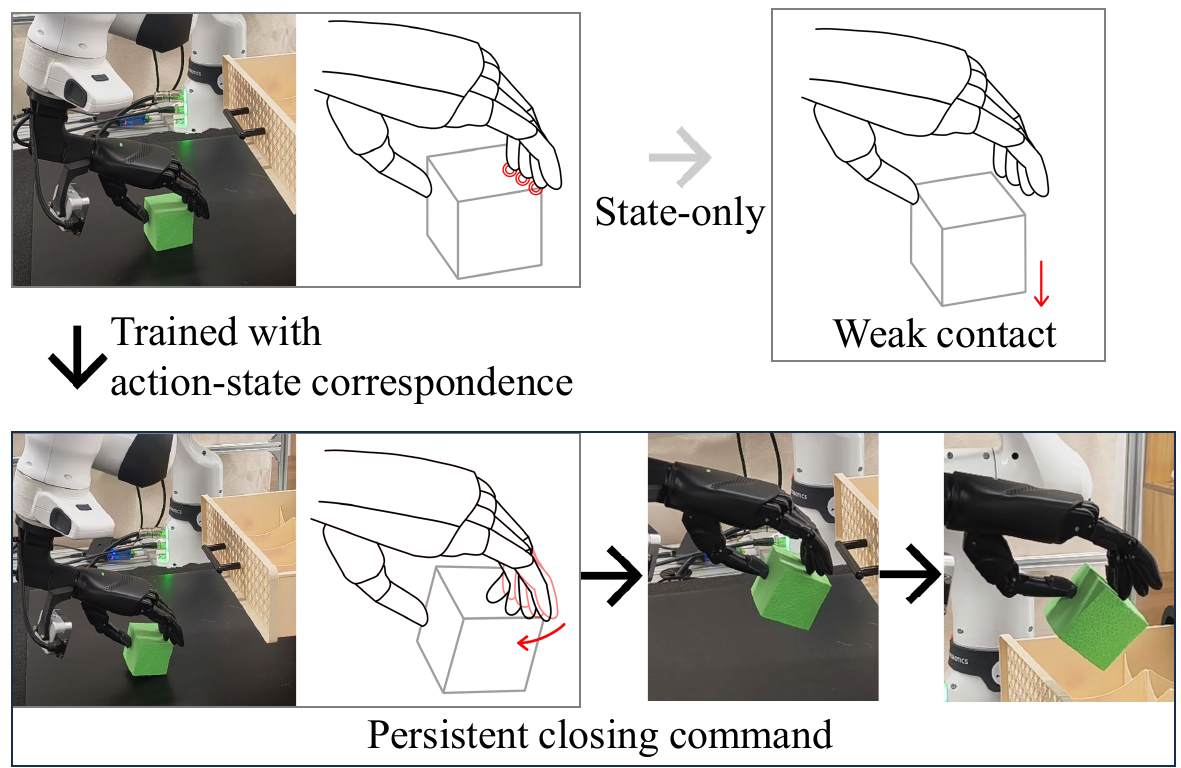}
\caption{
\textbf{Action--state correspondence.}
By learning from paired executable hand actions and states, the policy receives direct supervision for contact-aware corrections in contact-rich manipulation, which state-only supervision cannot provide.
}
  \label{fig:actionstate}
\end{figure}

Each episode records time-aligned hand commands, measured hand states, in-hand RGB observations, fingertip tactile signals, and 6-DoF tracker poses. The hand commands and sensory streams come from the same dexterous end-effector module used at deployment, with tracker poses used only to construct hand-frame relative motion labels. Thus, policy supervision uses the actual control inputs of the deployed hand, while contacts and visual--tactile observations retain the same fingertip surfaces and hand-relative geometry.

The resulting data provide hand action--state correspondence under contact, as illustrated in Fig.~\ref{fig:actionstate}. A measured hand state describes the posture reached after object contact constrains finger motion, whereas the paired command records the remaining control intent, such as continuing to close. Policies trained on these paired signals can associate visual and tactile contact cues with corrective finger commands, providing an implicit recovery signal that is absent from state-only labels.

Using this interface, we collect over 100 hours of demonstrations across more than eight dexterous tasks, with more than 200 episodes per task.
We evaluate policies on eight representative real-robot tasks and will release the collected dataset used in our experiments.


\section{Policy Learning and Deployment}
\label{sec:policy}

\subsection{Policy Interface}

From the recorded demonstration streams, the policy observation at timestep $t$ is
\begin{equation}
    o_t =
    \left(
    I_t,
    S_t^{\mathrm{tactile}},
    q_t^{\mathrm{hand}}
    \right),
    \label{eq:observation}
\end{equation}
where $I_t \in \mathbb{R}^{256 \times 256 \times 3}$ is the resized in-hand RGB image, $S_t^{\mathrm{tactile}} \in \mathbb{R}^{5 \times 10 \times 4}$ is the fingertip tactile signal, and $q_t^{\mathrm{hand}} \in \mathbb{R}^{6}$ is the actuated hand joint state.

The 6-DoF tracker is rigidly mounted to the hand module, and its pose is converted to a predefined hand reference frame through a fixed transform.
We denote the resulting hand-frame pose by \(T_t \in SE(3)\).
This absolute pose is not included in the policy observation.
Prior systems such as FastUMI~\citep{liu2025fastumi} and TouchGuide~\citep{zhang2026touchguide} map demonstration trajectories into robot execution frames through heuristic coordinate alignment.
This alignment hard-codes demonstrations to a predefined structured workspace and ties data collection to a deployment-time robot base frame.
Such an assumption is incompatible with robot-free collection in unstructured settings, where the collection frame may be unavailable or inconsistent with the robot base at deployment.
RealDexUMI instead uses \(T_t\) only to construct hand-frame relative action labels, so the policy predicts end-effector motion from local visual, tactile, and hand-state cues rather than memorizing a collection-specific global pose.


For each future step $t+k$ in an action chunk, we define
\begin{equation}
    \Delta T_{t,k} = T_t^{-1} T_{t+k}.
\end{equation}
We parameterize $\Delta T_{t,k}$ by translation $\Delta p_{t,k}$ and rotation vector $\Delta r_{t,k}$, and define the full action label as
\begin{equation}
    a_{t,k} =
    \left[
    \Delta p_{t,k},
    \Delta r_{t,k},
    u^{\mathrm{hand}}_{t+k}
    \right],
\end{equation}
where $u^{\mathrm{hand}}_{t+k}$ is the executable hand command captured from the isomorphic glove. 
This action space is expressed in the hand reference frame, decoupling policy supervision from the carrier robot while preserving the deployed hand's command interface.

\begin{figure}[H]
    \centering
    \includegraphics[
        width=\linewidth,
        trim=0mm 54.5mm 0mm 0mm,
        clip
    ]{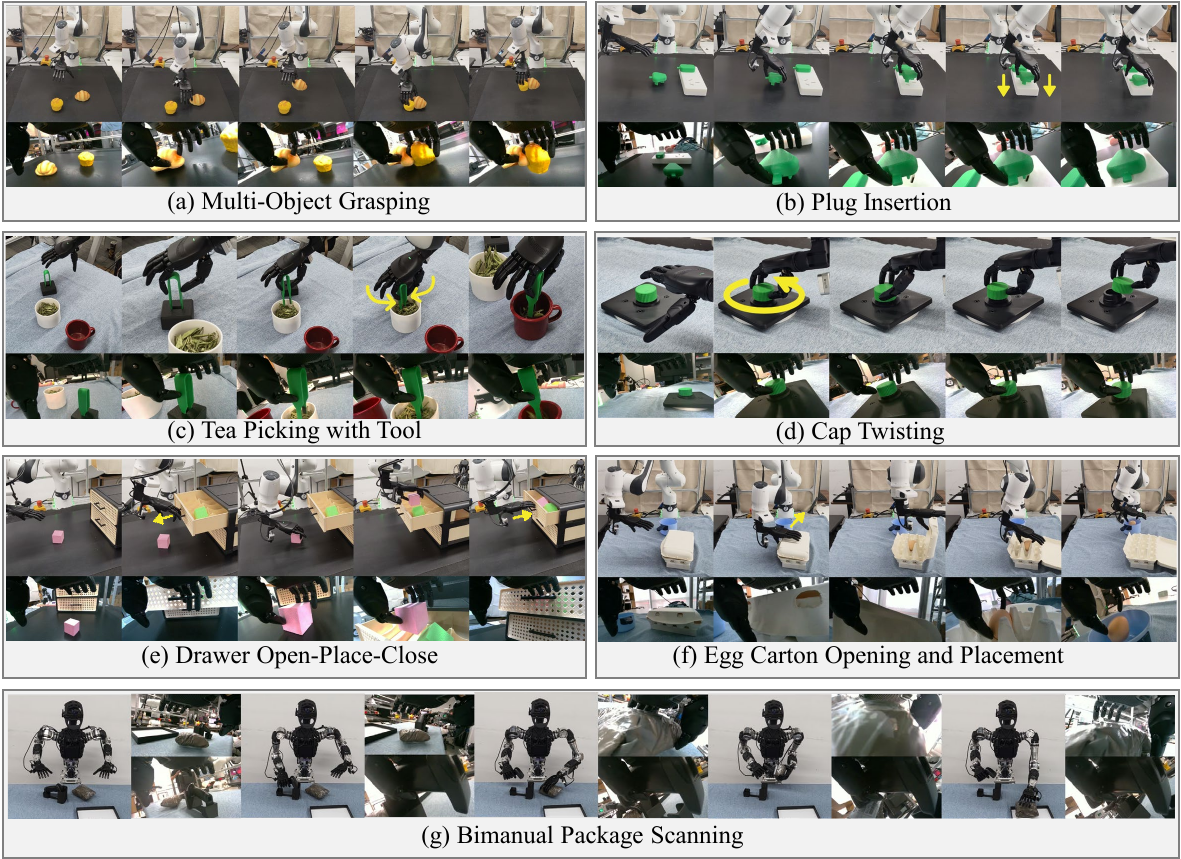}
\caption{
\textbf{Policy rollouts.}
Policies trained from RealDexUMI demonstrations execute representative tasks across multi-object grasping, precision insertion, tool use, twisting, articulated-object interaction, long-horizon execution, and bimanual operation.
}
    \label{fig:rollout}
\end{figure}

The policy predicts a chunk of future actions:
\begin{equation}
    \hat{A}_t =
    \pi_\theta(o_t)=
    \{\hat{a}_{t,1}, \ldots, \hat{a}_{t,C}\}.
\end{equation}
We instantiate $\pi_\theta$ with ACT~\cite{zhao2023learning} for all main experiments and train it on RealDexUMI demonstrations. 
The chunked prediction supervises hand-frame end-effector motion and finger commands as temporally coherent future sequences.
We additionally evaluate Diffusion Policy~\cite{chi2024diffusionpolicy} on selected tasks using the same observation and action representation.

\subsection{Deployment Interface}

RealDexUMI transfers across robot bodies by keeping the dexterous end-effector module and policy interface fixed while changing only the robot-side kinematics and controller.
At deployment, robot kinematics provide the current pose $\hat{T}_t \in SE(3)$ of the same hand reference frame.
Given a predicted relative action $\Delta \hat{T}_{t,k}$, the robot target is
\[
    \hat{T}^{\mathrm{target}}_{t,k}
    =
    \hat{T}_t \Delta \hat{T}_{t,k}.
\]
The robot-side controller realizes this target pose, while the shared dexterous hand directly executes the predicted hand command.
Thus, cross-embodiment deployment changes the IK and low-level controller, not the learned policy or dexterous end-effector interface.


\section{Experiments}
\label{sec:experiments}

We evaluate RealDexUMI as a wearable interface for deployable dexterous policy learning. 
Unless otherwise specified, policy-learning experiments use ACT trained from 200 RealDexUMI demonstrations per task and evaluated on a Franka FR3 equipped with the same dexterous end-effector module used during collection.
Each policy is evaluated over 20 real-robot trials, with success defined as completing the full task.

\subsection{Policies Learned from RealDexUMI Demonstrations}
\label{sec:policy_results}



\begin{figure}[H]
    \centering
    \captionsetup{font=small,skip=2pt}
    \includegraphics[
    width=1.0\linewidth,
    trim=0mm 0mm 0mm 0mm,
    clip
    ]{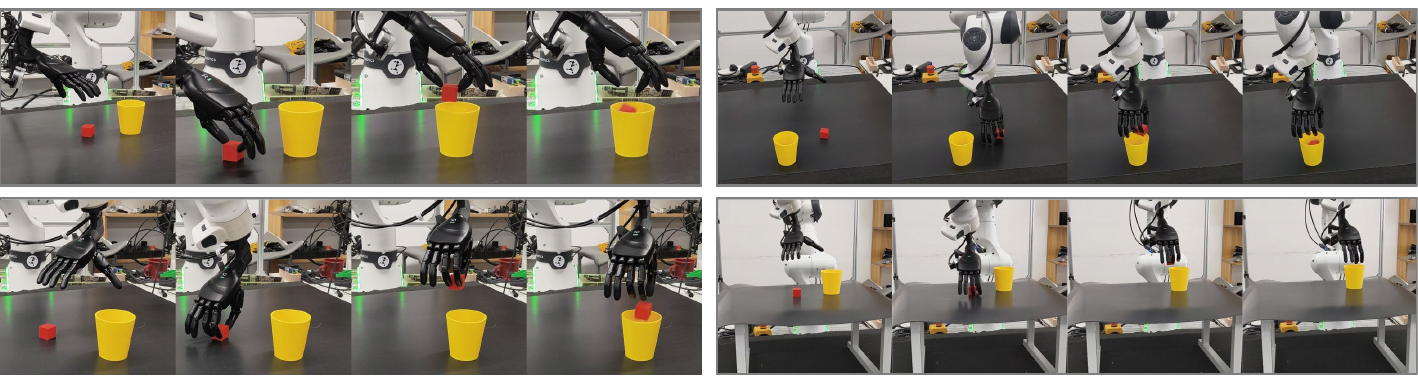}
    \caption{
    \textbf{Initial-pose robustness.}
    A cube pick-and-place policy is evaluated under unseen initial robot poses.
    }
    \label{fig:initpose}
\end{figure}



\paragraph{Initial-pose variation.}
Fig.~\ref{fig:initpose} evaluates the same cube pick-and-place checkpoint from left, right, center, and raised-center initial robot poses, with five trials per pose.
The policy succeeds in all 20 trials, suggesting that the hand-frame relative action representation helps the learned motion remain valid under changes in the robot's initial base-frame pose.

\begin{table}[H]
\centering
\small
\setlength{\tabcolsep}{2.5pt}
\renewcommand{\arraystretch}{1.05}
\caption{
Full-task success and single-factor ablations across eight real-robot tasks.
Per-task entries report success rates in $[0,1]$ over 20 trials, and Avg. reports the overall success percentage.
The w/o tactile variant keeps executable hand-command labels but removes tactile observations.
State-as-action keeps tactile observations but uses the next measured hand state as the hand-action label and executes the predicted state as the hand command.
}
\label{tab:table_experiment}
\begin{tabular}{
L{2.3cm}
C{1.0cm}
C{1.4cm}
C{1.0cm}
C{1.0cm}
C{1.0cm}
C{1.0cm}
C{1.0cm}
C{1.0cm}
C{1.5cm}
}
\toprule
\textbf{Method} 
& \textbf{Cube} 
& \textbf{Multi-obj.} 
& \textbf{Plug} 
& \textbf{Cap} 
& \textbf{Tea} 
& \textbf{Drawer} 
& \textbf{Egg} 
& \textbf{Biman.} 
& \textbf{Avg.} \\
\midrule
\textbf{RealDexUMI} 
& 1.00 
& 1.00 
& 0.85 
& 1.00 
& 0.80 
& 0.85 
& 0.70 
& 0.90 
& \textbf{88.75\%} \\
w/o tactile 
& 0.90 
& 1.00 
& 0.45 
& 0.80 
& 0.55 
& 0.70 
& 0.60 
& 0.60 
& 70.00\% \\
State-as-action 
& 0.65 
& 0.85 
& 0.30 
& 0.35 
& 0.20 
& 0.45 
& 0.60 
& 0.70 
& 51.25\% \\
\bottomrule
\end{tabular}
\end{table}

\paragraph{Policy performance.}
RealDexUMI achieves an average full-task success rate of 88.75\% across eight real-robot tasks.
Together with the rollouts in Fig.~\ref{fig:rollout}, this shows that RealDexUMI demonstrations preserve deployable dexterity across contact-rich, tool-use, articulated, long-horizon, and bimanual manipulation.
For multi-stage tasks, Table~\ref{tab:table_experiment} reports only final task completion, while Appendix~Table~\ref{tab:subgoal_completion} provides cumulative subgoal completion for failure analysis.

\paragraph{Ablation analysis.}
Removing tactile input reduces average success from 88.75\% to 70.00\%, mainly on tasks where contact is difficult to infer from vision alone, such as plug insertion and tea picking.
State-as-action supervision further reduces success to 51.25\%, especially when policies must maintain or recover contact, supporting the importance of action--state correspondence and executable hand-command supervision.

\begin{figure}[H]
    \centering
    \includegraphics[
        width=\linewidth,
        trim=0mm 0mm 0mm 0mm,
        clip
    ]{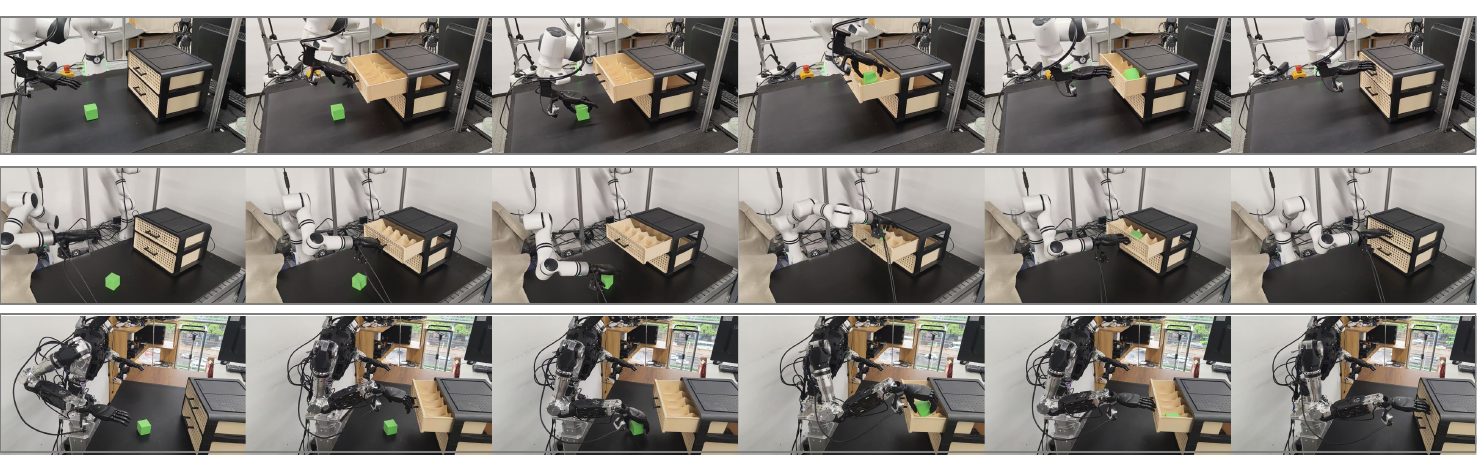}
\caption{
\textbf{Cross-embodiment deployment.}
The same drawer-stowing checkpoint runs on Franka FR3, RealMan RM65, and PND Adam-U without retraining.
}
    \label{fig:cross}
\end{figure}



\subsection{Cross-Embodiment Deployment}
\label{sec:cross_embodiment_exp}

\begin{table}[H]
    \centering
    \captionsetup{font=small,skip=2pt}
    \caption{
    \textbf{Cross-embodiment success.}
    }
    \label{tab:cross_embodiment}
    \small
    \renewcommand{\arraystretch}{1.02}
    \begin{tabular}{lccc}
        \toprule
        \textbf{Task} & \textbf{FR3} & \textbf{RM65} & \textbf{Adam-U} \\
        \midrule
        Cube   & 1.00 & 1.00 & 1.00 \\
        Drawer & 0.85 & 0.75 & 0.80 \\
        \bottomrule
    \end{tabular}
\end{table}

RealDexUMI decouples dexterous skill learning from the carrier robot by predicting hand-frame motions and hand commands, leaving each robot to realize them with its own IK and low-level controller.
We evaluate the same policy checkpoints on a 7-DoF Franka FR3, a 6-DoF RealMan RM65, and one 7-DoF arm of the dual-arm PND Adam-U. 
Each task--embodiment pair is evaluated over 20 real-robot trials, for a total of 120 trials.
As shown in Table~\ref{tab:cross_embodiment} and Fig.~\ref{fig:cross}, the checkpoints deploy without retraining across all three embodiments and achieve high success rates on both tasks.
These results indicate that the learned dexterous behavior is tied primarily to the shared end-effector interface rather than to a specific robot arm.



\subsection{Collection-Time Dexterity}

\begin{figure}[htbp]
    \centering
    \includegraphics[
    width=0.5\linewidth,
    trim=0.5mm 0mm 0mm 0.5mm,
    clip
    ]{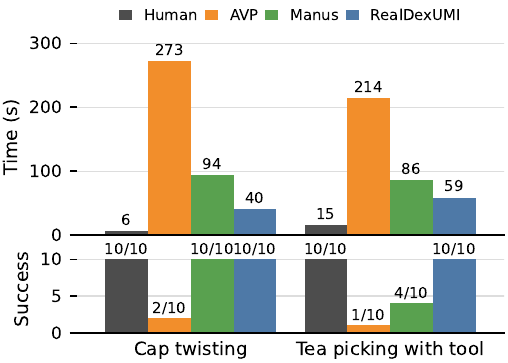}
\caption{
\textbf{Teleoperation comparison.}
Time is averaged over successful trials.
Trials exceeding 5 min are counted as failures.
}
    \label{fig:teleop_efficiency}
\end{figure}

We evaluate collection-time control on two dexterous tasks: cap twisting and tea picking with tweezers.
We compare RealDexUMI with AVP-based arm--hand teleoperation~\citep{park2024avp} and Manus-glove retargeting to the RealDexUMI~\citep{xin2026analyzing}, and include direct human-hand manipulation as a reference.

As shown in Fig.~\ref{fig:teleop_efficiency}, success rate is the primary metric, since completion time is averaged only over successful trials, therefore it measures efficiency conditional on task completion.
RealDexUMI achieves the highest success rate and shortest completion time among deployable collection interfaces.

The Manus baseline succeeds on cap twisting, suggesting that wearable hand input can make wrist-driven teleoperation intuitive.
However, its performance drops sharply on tea picking, where tweezer use depends on precise fingertip pinching.
This contrast shows that rich human-hand motion alone is insufficient when it must be retargeted to a robot hand.
RealDexUMI instead provides precise linear control directly in the dexterous hand's action space, enabling reliable deployable demonstrations.


\section{Limitations}
\label{sec:limitation}

RealDexUMI prioritizes end-effector alignment, so its current sensing is mainly local.
This limits tasks that require object search, long-range planning, or explicit task-progress reasoning.
Adding egocentric or global views is a promising direction, but keeping such views aligned between wearable collection and robot deployment remains challenging.

RealDexUMI also reflects a trade-off between dexterity, weight, and wearable controllability.
The current hand with six active DoFs keeps the system lightweight and enables intuitive isomorphic glove control, but it does not cover the full capability of higher-DoF dexterous hands.
Extending this paradigm to more expressive hands while preserving low-burden, precise, and deployment-aligned control remains future work.

\section{Conclusion}
\label{sec:conclusion}

We presented RealDexUMI, a wearable interface for collecting deployable dexterous manipulation data. 
RealDexUMI makes the deployable dexterous hand itself the demonstration interface, and uses a palm-side isomorphic glove to provide precise, low-burden control in the hand's executable action space.
This design removes the need to teleoperate a robot body during collection while preserving the end-effector behavior needed for deployment.
Policies trained from RealDexUMI data achieve 88.75\% average success across eight real-world dexterous tasks and transfer across heterogeneous embodiments by replacing only the IK and low-level controller.
These results highlight RealDexUMI as a practical and scalable interface for robot-free collection of deployable dexterous manipulation data.

\clearpage
\acknowledgments{If a paper is accepted, the final camera-ready version will (and probably should) include acknowledgments. All acknowledgments go at the end of the paper, including thanks to reviewers who gave useful comments, to colleagues who contributed to the ideas, and to funding agencies and corporate sponsors that provided financial support.}


\bibliographystyle{unsrtnat}
\bibliography{citation}  

\clearpage
\appendix
\section*{Appendix}

\addcontentsline{toc}{section}{Appendix}

\section{Glove Hardware and Embedded Interface}

\begin{figure}[H]
	    \centering
	    \includegraphics[
	    width=0.5\linewidth,
	    trim=0.5mm 0mm 0mm 0.5mm,
	    clip
	    ]{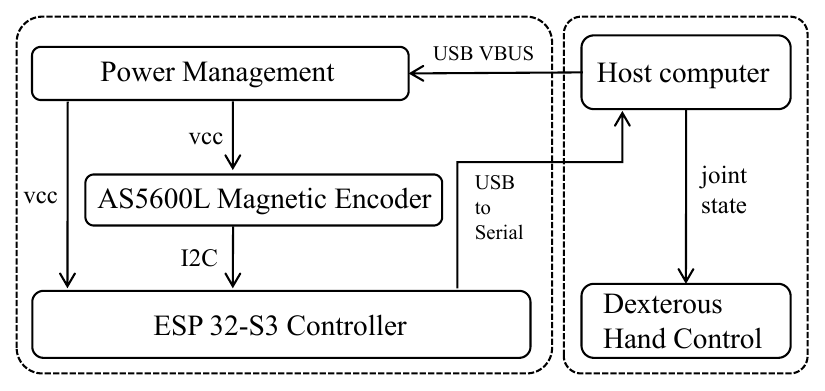}
    \caption{
	    \textbf{Glove embedded interface.}
	    Six AS5600L magnetic encoders measure the actuated glove DoFs.
	    The ESP32-S3 controller reads encoder values through I2C and streams the resulting 6-D command vector to the host computer through USB serial.
		    \label{fig:embedded_system_overview}
		    }
\end{figure}

This section provides implementation details of the palm-side teleoperation glove used for RealDexUMI data collection.
We focus on how the glove measures operator inputs and converts them into executable hand commands.
Fig.~\ref{fig:embedded_system_overview} shows the overall embedded interface of the glove.

\subsection{Glove Sensing Interface}

The glove measures six actuated command DoFs using magnetic encoders.
Each measured DoF corresponds to one actuated DoF of the RealDexUMI hand, producing a 6-D command vector in the deployed hand's command space.
Unlike human-hand motion capture, this interface does not estimate fingertip poses or solve a retargeting problem.
It directly records the command variables later executed by the robot hand.

\begin{figure}[H]
\centering
\includegraphics[width=0.85\linewidth]{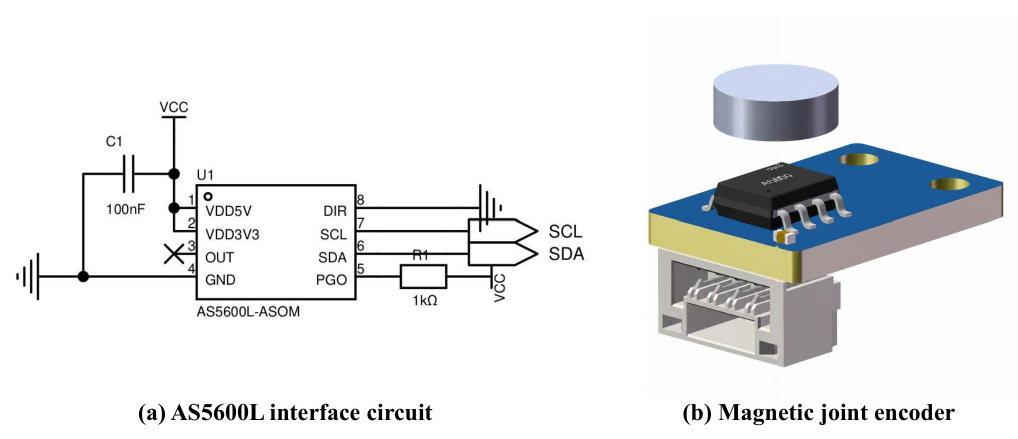}
\caption{
\textbf{Single-joint magnetic encoder design.}
(a) AS5600L interface circuit. The encoder communicates with the controller through the I2C bus using SCL and SDA.
(b) Magnetic joint encoder. A diametric magnet is aligned with the joint rotation axis and placed above the AS5600L sensor for absolute angle measurement.
}
\label{fig:magnetic_joint_encoder}
\end{figure}

Fig.~\ref{fig:magnetic_joint_encoder} illustrates the single-joint encoder design.
For each sensed DoF, a diametric magnet is aligned with the joint rotation axis and placed above an AS5600L sensor.
The sensor provides an absolute angular reading by measuring the magnetic field direction of the rotating magnet.
This non-contact measurement avoids mechanical friction in the sensing path and allows the glove to recover the current joint reading after power-on.

\subsection{Embedded Controller}

An ESP32-S3 controller reads the six encoder values through the I2C bus and streams the measurements to the host computer through a USB serial interface, as shown in Fig.~\ref{fig:embedded_system_overview}.
The controller packages the encoder readings into a fixed-order joint vector, which is then used by the host-side teleoperation process to command the dexterous hand.
This design keeps the glove self-contained and allows it to be used without an external data-acquisition board.

\subsection{Command Calibration and Mapping}

Before each collection session, the glove is held at a reference open pose to record encoder offsets.
During operation, each encoder reading is converted into a relative displacement from this reference pose and linearly mapped to the corresponding hand command range.
The resulting command vector is recorded as the hand action label in the demonstration dataset and sent to the hand during teleoperation.

\section{Data Collection and Synchronization}

RealDexUMI records all demonstration streams at the end effector, including in-hand RGB observations, fingertip tactile signals, measured hand states, glove commands, and 6-DoF end-effector poses.
The RGB, tactile, hand-state, and hand-command streams are generated by the same dexterous end-effector module used during robot deployment.
The 6-DoF pose stream is used to construct relative end-effector action labels, but is not included in the policy observation.

\subsection{Recorded Streams}

Table~\ref{tab:recorded_streams} summarizes the recorded streams and their use in policy learning.
The policy observation consists of in-hand RGB, tactile signals, and measured hand states.
The action label consists of hand-frame relative end-effector motion and executable hand commands.

\begin{table}[h]
\centering
\caption{Recorded streams in RealDexUMI demonstrations.}
\label{tab:recorded_streams}
\footnotesize
\renewcommand{\arraystretch}{1.12}
\setlength{\tabcolsep}{5pt}
\begin{tabular}{lccc}
\toprule
\textbf{Stream} & \textbf{Typical rate} & \textbf{Policy observation} & \textbf{Action label} \\
\midrule
In-hand RGB & 30 Hz & Yes & No \\
Fingertip tactile & 20 Hz & Yes & No \\
Hand state & 100 Hz & Yes & No \\
Glove command & 200 Hz & No & Yes \\
6-DoF end-effector pose & 100 Hz & No & Yes \\
\bottomrule
\end{tabular}
\end{table}

\subsection{Latest-Sample Synchronization}

The recorded streams run at different rates, while policy training uses fixed-rate demonstration steps.
For each policy timestep, we use the in-hand RGB timestamp as the temporal anchor.
For every non-visual stream, we select the latest available sample whose timestamp is not later than the anchor timestamp.
This latest-sample strategy avoids extrapolating sensor values and keeps each training sample grounded in measurements that were already available at that time.

The same synchronized timestep provides both policy observations and action labels.
The observation is formed from the RGB frame, the latest tactile signal, and the latest measured hand state.
For each future action step $t+k$, the hand-command label is selected from the latest glove command not later than the corresponding future timestamp. 
The end-effector motion label is computed from the synchronized 6-DoF pose sequence as a relative transform in the current hand frame.
Thus, the policy learns from local end-effector observations and predicts actions in the same hand-centric interface used during deployment.

\section{Task Definitions and Evaluation Protocol}

The task suite covers basic grasping, rotational manipulation, articulated-object interaction, tool use, multi-stage manipulation, precision insertion, multi-object grasping, and bimanual coordination.
Each policy is evaluated over 20 real-robot trials per task.
A trial is counted as successful only if the final task goal is completed within the task timeout.
Table~\ref{tab:task_definitions} summarizes the full-task success criterion for each task.

\begin{table}[t]
\centering
\caption{Task definitions and full-task success criteria.}
\label{tab:task_definitions}
\footnotesize
\renewcommand{\arraystretch}{1.08}
\setlength{\tabcolsep}{4pt}
\begin{tabular}{L{3.2cm}L{9.8cm}}
\toprule
\textbf{Task} & \textbf{Full-task success criterion} \\
\midrule
Cube pick-and-place
& The cube is placed into the target cup. \\

Cap twisting
& The lid is twisted off from the fixture. \\

Drawer stowing
& The drawer is opened, the cube is placed inside, and the drawer is closed. \\

Tea picking with tool
& The tweezers are grasped and used to transfer tea into the target cup. \\

Egg carton handling
& The carton is opened, the egg is picked, and the egg is placed into the pot. \\

Plug insertion
& The plug is inserted into the target socket. \\

Multi-object grasping
& Both target objects are grasped in sequence. \\

Package scanning
& The package is manipulated bimanually and scanned successfully. \\
\bottomrule
\end{tabular}
\end{table}

\subsection{Evaluation Protocol}

For each task, the initial object and robot configurations are varied within the feasible workspace of the robot.
All trials are evaluated on physical robot rollouts without resetting the policy during execution.
For long-horizon tasks, intermediate subgoals are recorded in the order required by the task.
A later subgoal is counted only when all preceding subgoals in the same rollout have been completed.

This cumulative subgoal protocol is used only to diagnose failure modes.
We do not average subgoal success rates as independent trials, because later subgoals are conditioned on earlier task progress.
The final task success reported in the main paper is therefore the primary metric.

\subsection{Representative Task Props}

\begin{figure}[H]
\centering

\hspace*{0.07\linewidth}
\begin{minipage}[t]{0.36\linewidth}
\centering
\includegraphics[
    width=\linewidth,
    trim=40mm 37mm 40mm 40mm,
    clip
]{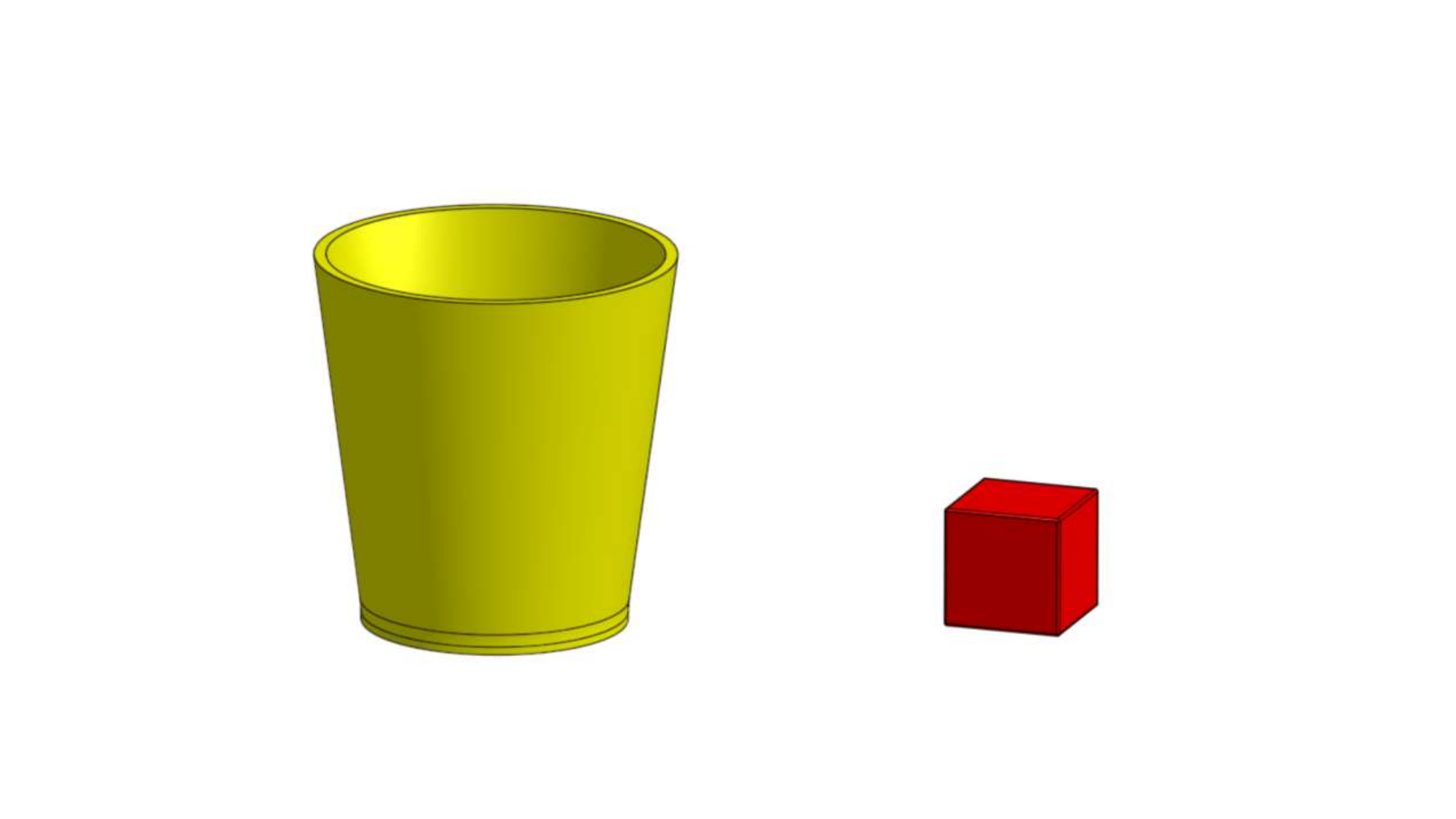}

{\small (a) Cube pick-and-place.}
\end{minipage}
\hfill
\begin{minipage}[t]{0.36\linewidth}
\centering
\includegraphics[
    width=\linewidth,
    trim=40mm 37mm 40mm 40mm,
    clip
]{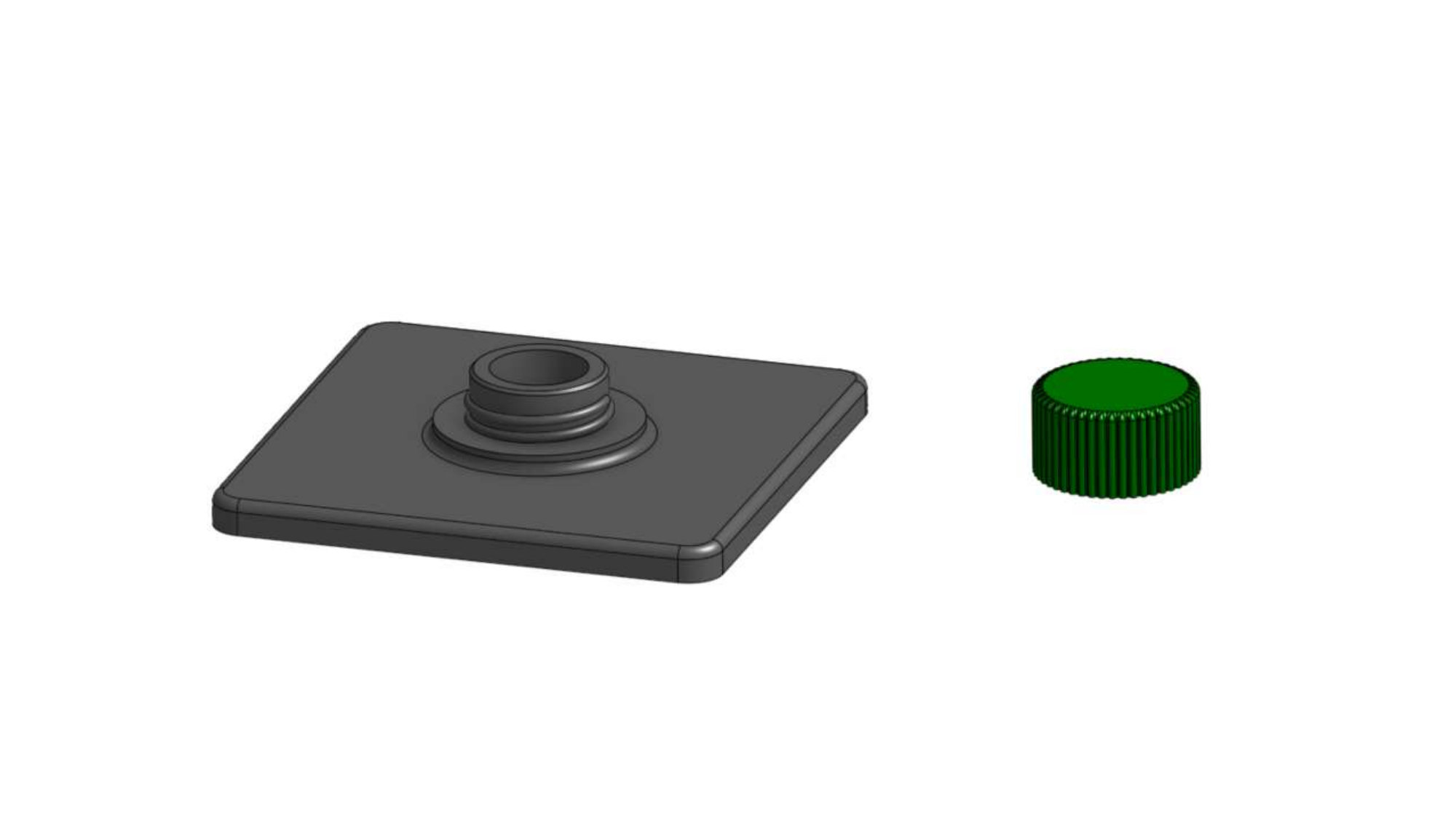}

{\small (b) Cap twisting.}
\end{minipage}
\hspace*{0.09\linewidth}

\vspace{2mm}

\hspace*{0.07\linewidth}
\begin{minipage}[t]{0.36\linewidth}
\centering
\includegraphics[
    width=\linewidth,
    trim=40mm 37mm 40mm 40mm,
    clip
]{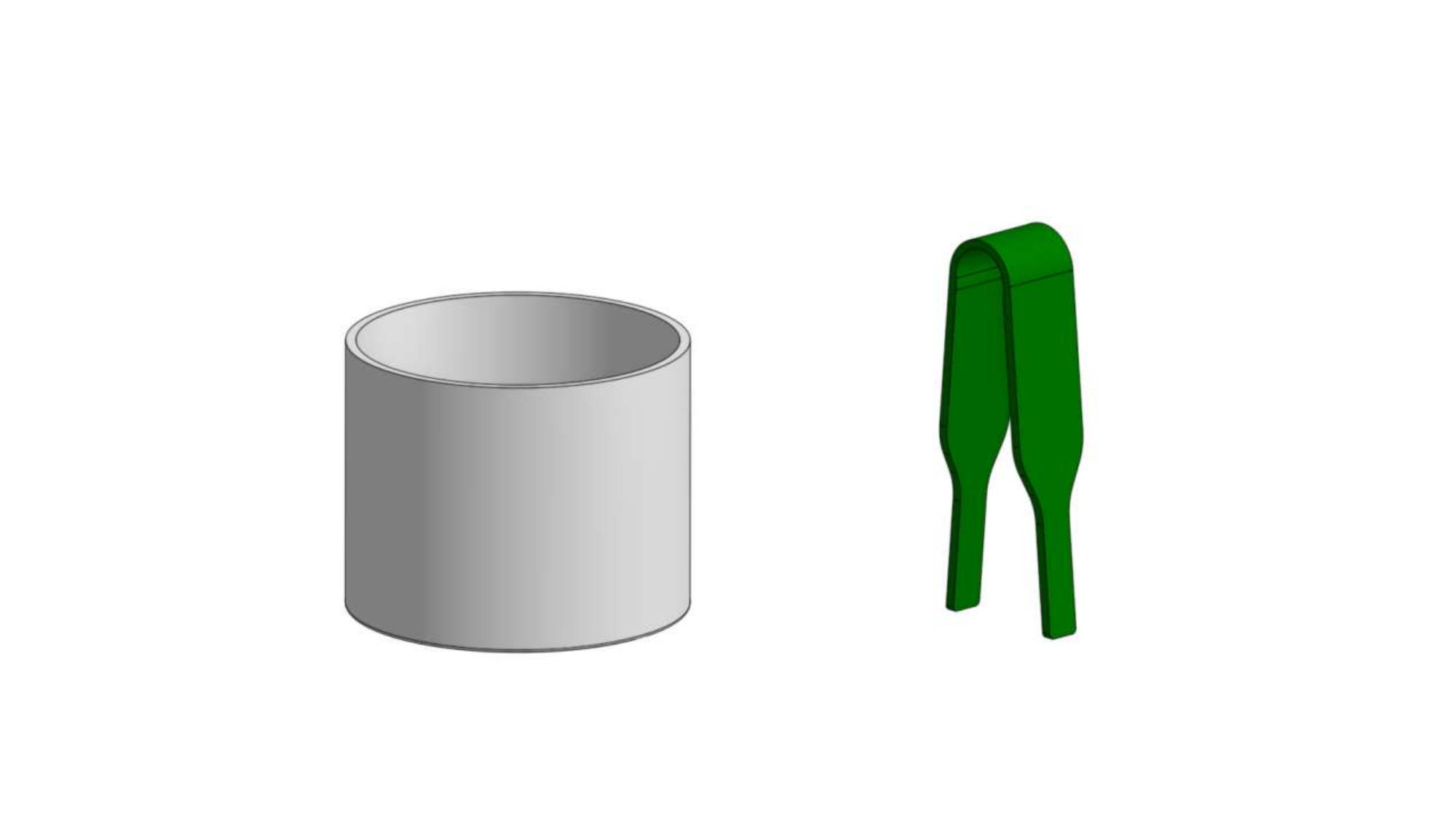}

{\small (c) Tea picking with tool.}
\end{minipage}
\hfill
\begin{minipage}[t]{0.36\linewidth}
\centering
\includegraphics[
    width=\linewidth,
    trim=40mm 37mm 40mm 40mm,
    clip
]{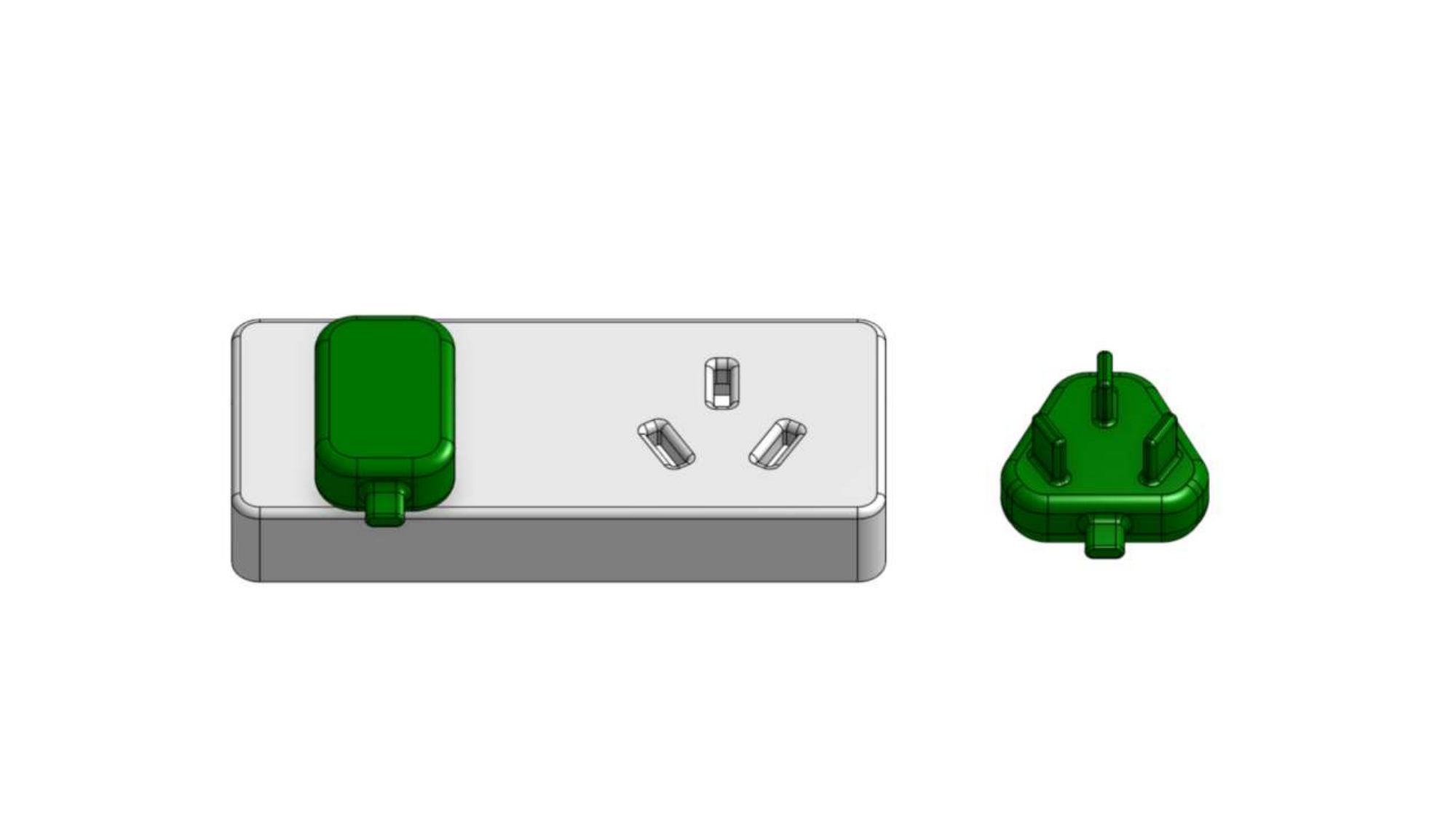}

{\small (d) Plug insertion.}
\end{minipage}
\hspace*{0.09\linewidth}

\caption{
\textbf{Representative task props.}
We use simple fixtures or 3D-printed props for tasks where controlled geometry is useful:
a cup and 2.5 cm cube for cube pick-and-place,
a ridged cap fixture for lid twisting,
a tweezer-and-cup setup for tea picking,
and a plug fixture for insertion.
}
\label{fig:representative_task_props}
\end{figure}

Several tasks use simple fixtures or 3D-printed props to standardize task geometry and improve reproducibility.
These props are used where controlled geometry is useful, but not every task requires a printed asset.
Fig.~\ref{fig:representative_task_props} shows representative props used in the cube pick-and-place, lid twisting, tea picking, and plug insertion.

\section{Training Details}

\begin{figure}[H]
    \centering
    \includegraphics[
        width=1.0\linewidth,
        trim=0mm 0mm 0mm 0mm,
        clip
    ]{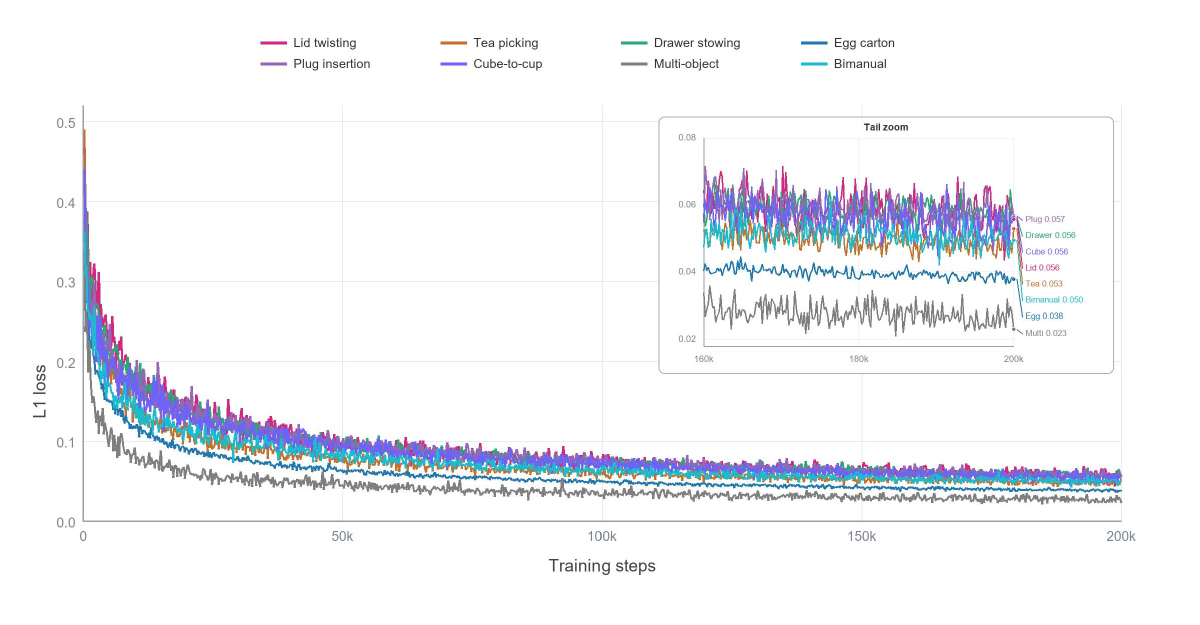}
    \caption{
    \textbf{Policy training loss.}
    Action-prediction L1 loss during training.
    }
    \label{fig:l1_loss}
\end{figure}

We train ACT for the main real-robot policy experiments and evaluate Diffusion Policy as an additional imitation-learning baseline.
Both policy families use the same RealDexUMI observation and action interface.
The observation consists of in-hand RGB, fingertip tactile signals, and measured hand states.
The action consists of hand-frame relative end-effector motion and executable hand commands.
Table~\ref{tab:training_hyperparameters} summarizes the training hyperparameters used for ACT and Diffusion Policy.
Fig.~\ref{fig:l1_loss} reports the action-prediction L1 loss during policy training.
We include this plot as an optimization diagnostic, real-robot success rates remain the primary evaluation metric.

\begin{table}[htb]
\centering
\caption{Training hyperparameters for ACT and Diffusion Policy.}
\label{tab:training_hyperparameters}
\renewcommand{\arraystretch}{1.15}
\resizebox{\linewidth}{!}{
\begin{tabular}{llcc}
\toprule
\textbf{Group} & \textbf{Parameter} & \textbf{ACT} & \textbf{Diffusion Policy} \\
\midrule
Observation & Image obs. horizon & 1 & 2 \\
& Proprio. obs. horizon & 1 & 2 \\
& Action horizon & 20 & 32 \\
& Executed action steps & 20 & 32 \\
& Obs. resolution & $256\times256$ & $256\times256$ \\
\midrule
Architecture & Vision encoder & ResNet-18 & ResNet-18 \\
& Pretrained vision weights & ImageNet & None \\
& Transformer dim. & 512 & -- \\
& Transformer layers & 4 encoder, 1 decoder & -- \\
& Attention heads & 8 & -- \\
& ACT latent dim. & 32 & -- \\
& KL weight & 10 & -- \\
& U-Net channels & -- & $(512,1024,2048)$ \\
\midrule
Optimization & Optimizer & AdamW & Adam \\
& Learning rate & $1\times10^{-5}$ & $1\times10^{-4}$ \\
& Momentum & $\beta_1=0.9,\ \beta_2=0.999$ & $\beta_1=0.95,\ \beta_2=0.999$ \\
& Weight decay & $1\times10^{-4}$ & $1\times10^{-6}$ \\
& LR schedule & None & Cosine, 500 warmup steps \\
& Batch size & 128 & 128 \\
& Training steps & 200k & 100k \\
& Grad. clip norm & 10 & 10 \\
\midrule
Diffusion & Train diffusion steps & -- & 100 \\
& Inference steps & -- & 100 \\
& Noise scheduler & -- & DDPM \\
& Beta schedule & -- & Squared cosine \\
& Prediction target & -- & Noise $\epsilon$ \\
\bottomrule
\end{tabular}
}
\end{table}

\subsection{Diffusion Policy Baseline}
\label{app:dp_baseline}

We additionally train Diffusion Policy on the same RealDexUMI demonstrations using the same observation and action interface as the ACT policies in the main experiments.
This evaluation tests whether RealDexUMI data can support another common imitation-learning policy backend.
Table~\ref{tab:dp_success} reports the Diffusion Policy results across the same eight real-robot tasks.

\begin{table}[H]
\centering
\small
\setlength{\tabcolsep}{3pt}
\renewcommand{\arraystretch}{1.05}
\caption{
Diffusion Policy success rates across eight real-robot tasks.
Per-task entries report success rates in $[0,1]$ over 20 trials, and Avg. reports the overall success percentage.
}
\label{tab:dp_success}
\begin{tabular}{
L{2.3cm}
C{1.0cm}
C{1.4cm}
C{1.0cm}
C{1.0cm}
C{1.0cm}
C{1.0cm}
C{1.0cm}
C{1.0cm}
C{1.5cm}
}
\toprule
\textbf{Method} 
& \textbf{Cube} 
& \textbf{Multi-obj.} 
& \textbf{Plug} 
& \textbf{Cap} 
& \textbf{Tea} 
& \textbf{Drawer} 
& \textbf{Egg} 
& \textbf{Biman.} 
& \textbf{Avg.} \\
\midrule
Diffusion Policy
& 0.85
& 0.75
& 0.25
& 0.75
& 0.40
& 0.60
& 0.80
& 0.70
& 63.75\% \\
\bottomrule
\end{tabular}
\end{table}

The Diffusion Policy results are lower than the ACT results reported in the main text, especially on precision- and contact-rich tasks.
We use this experiment to test backend compatibility rather than to exhaustively tune policy architectures, and therefore use ACT as the primary backend in the main experiments.

\section{Additional Evaluation Results}

\subsection{Cumulative Subgoal Completion}


Table~\ref{tab:subgoal_completion} reports cumulative subgoal completion for multi-stage tasks, showing where failures occur.
A later subgoal is counted only if all preceding subgoals in the same rollout have been completed.
Therefore, these values should not be interpreted as independent subgoal success rates; the final full-task success remains the primary metric reported in the main paper.

\begin{table}[t]
\centering
\caption{Cumulative subgoal completion for multi-stage tasks.}
\label{tab:subgoal_completion}
\renewcommand{\arraystretch}{1.12}
\setlength{\tabcolsep}{6pt}
\begin{tabular}{lclc}
\toprule
\textbf{Task} & \textbf{Step} & \textbf{Subgoal} & \textbf{Cumulative completion} \\
\midrule
\multirow[c]{3}{*}{Drawer stowing}
& 1 & Open drawer & 20/20 \\
& 2 & Pick cube & 18/20 \\
& 3 & Close drawer & 17/20 \\
\midrule
\multirow[c]{3}{*}{Tea picking with tool}
& 1 & Grasp tweezers & 17/20 \\
& 2 & Pick tea & 16/20 \\
& 3 & Place tea in cup & 16/20 \\
\midrule
\multirow[c]{3}{*}{Egg carton handling}
& 1 & Open carton & 15/20 \\
& 2 & Pick egg & 14/20 \\
& 3 & Place egg in pot & 14/20 \\
\midrule
\multirow[c]{2}{*}{Multi-object grasping}
& 1 & Grasp first object & 20/20 \\
& 2 & Grasp second object & 20/20 \\
\bottomrule
\end{tabular}
\end{table}

\section{Collection-Time Interface Comparison}

We compare RealDexUMI with AVP-based arm--hand teleoperation and Manus-glove retargeting on two dexterous collection tasks.
The same operator performs all methods and receives 10 minutes of practice for each interface before evaluation.
For AVP, wrist and hand information are captured and retargeted to the Franka arm and RealDexUMI end-effector module.
For Manus-glove retargeting, the glove measurements are mapped to the RealDexUMI hand for collection-time control.
Completion time is averaged over successful trials, and trials that exceed the time limit are counted as failures.

\clearpage
\section{Survey for Perceived Teleoperation Complexity}

To support the perceived-complexity label in Table~\ref{tab:system_comparison}, we conduct a lightweight structured survey using available system descriptions, device images, and demonstration materials.
The rating reflects perceived setup and operation complexity and should be interpreted as a coarse comparative indicator rather than a direct hands-on usability measurement.

\begin{figure}[H]
\centering
\begingroup
\setlength{\fboxsep}{6pt}
\setlength{\fboxrule}{0.5pt}
\fbox{%
\begin{minipage}{0.96\textwidth}
\centering

\includegraphics[width=0.97\linewidth]{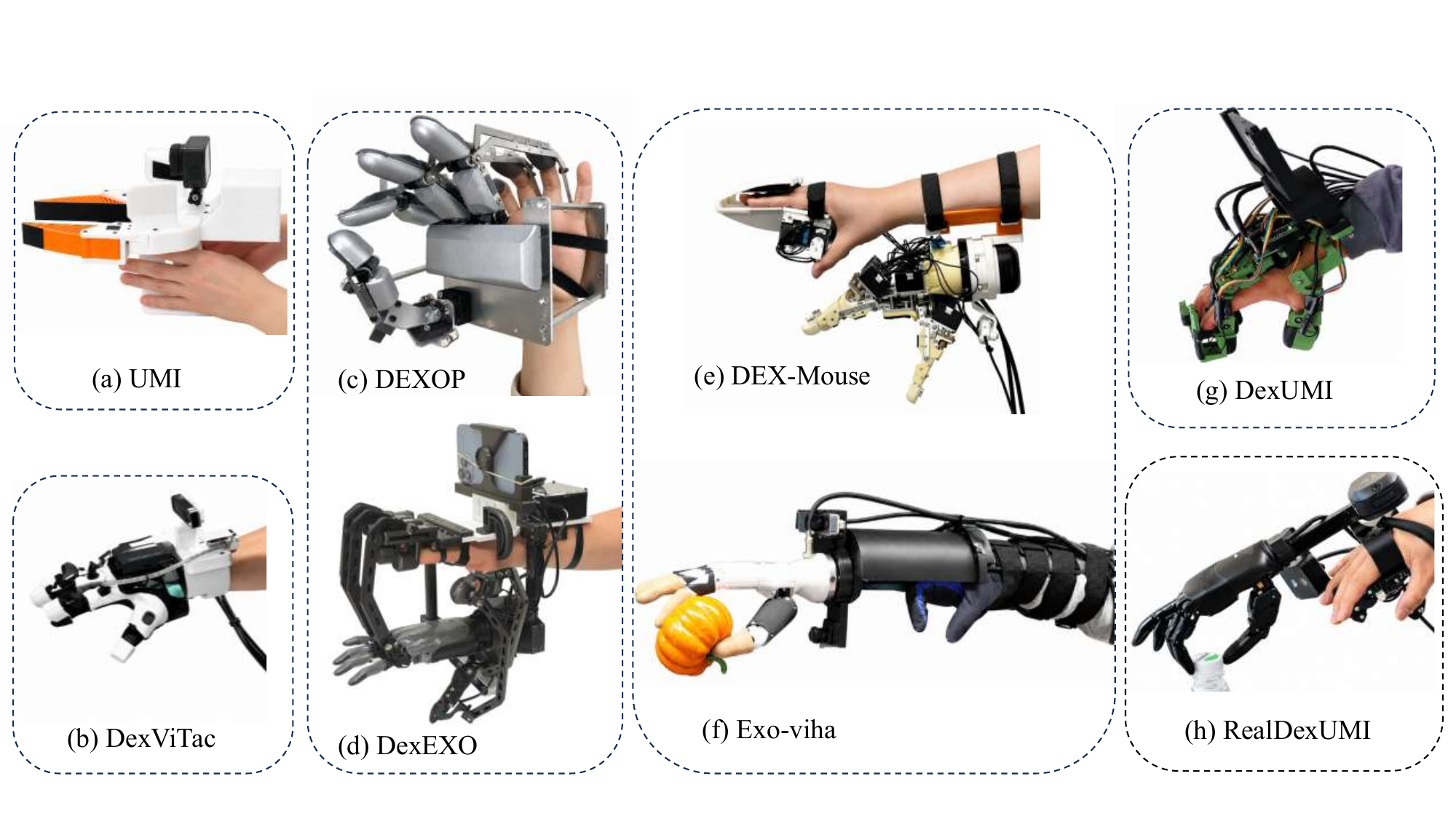}

\renewcommand{\arraystretch}{1.25}
\setlength{\tabcolsep}{4pt}

\begin{tabular*}{\linewidth}{
@{}
p{0.40\linewidth}
@{\extracolsep{\fill}}
>{\centering\arraybackslash}p{0.16\linewidth}
>{\centering\arraybackslash}p{0.16\linewidth}
>{\centering\arraybackslash}p{0.16\linewidth}
@{}
}
\toprule
\multicolumn{4}{p{\linewidth}}{
\textbf{Survey question:} How complex does the system appear for teleoperation-based demonstration collection, considering wearing/holding, setup, adjustment, and operation?
} \\
\midrule
\textbf{System} & \textbf{Low} & \textbf{Medium} & \textbf{High} \\
\midrule
(a)UMI~\citep{chi2024universal} & $\square$ & $\square$ & $\square$ \\
(b)DexViTac~\citep{chen2026dexvitac} & $\square$ & $\square$ & $\square$ \\
(c)DEXOP~\citep{fang2025dexop} & $\square$ & $\square$ & $\square$ \\
(d)DexExo~\citep{zhu2026dexexo} & $\square$ & $\square$ & $\square$ \\
(e)DEX-Mouse~\citep{koh2026dex} & $\square$ & $\square$ & $\square$ \\
(f)Exo-ViHa~\citep{chao2025exo} & $\square$ & $\square$ & $\square$ \\
(g)DexUMI~\citep{xu2025dexumi} & $\square$ & $\square$ & $\square$ \\
(h)RealDexUMI & $\square$ & $\square$ & $\square$ \\
\bottomrule
\end{tabular*}

\end{minipage}%
}
\endgroup
\caption{
\textbf{Survey form for perceived teleoperation complexity.}
Evaluators rate the perceived setup and operation complexity of each demonstration interface using a three-level scale.
}
\label{fig:survey_form}
\end{figure}

Evaluators rate each system on a three-level scale: Low, Medium, or High.
The final label is determined by majority vote across 20 evaluators with backgrounds in robotics or related engineering fields.
Fig.~\ref{fig:survey_form} shows the survey form provided to evaluators.
Table~\ref{tab:teleop_complexity_survey} reports the vote counts and final perceived-complexity labels.

\begin{table}[t]
\centering
\caption{
Survey results for perceived teleoperation complexity.
Each entry reports the number of votes among 20 evaluators.
}
\label{tab:teleop_complexity_survey}
\renewcommand{\arraystretch}{1.12}
\setlength{\tabcolsep}{5pt}
\begin{tabular}{lcccc}
\toprule
\textbf{System} & \textbf{Low} & \textbf{Medium} & \textbf{High} & \textbf{Final} \\
\midrule
UMI~\citep{chi2024universal}        & 18 & 2  & 0  & Low \\
DexViTac~\citep{chen2026dexvitac}   & 15 & 5  & 0  & Low \\
DEXOP~\citep{fang2025dexop}         & 0  & 0  & 20 & High \\
DexExo~\citep{zhu2026dexexo}        & 1  & 6  & 13 & High \\
DEX-Mouse~\citep{koh2026dex}        & 1  & 5  & 14 & High \\
Exo-ViHa~\citep{chao2025exo}        & 0  & 1  & 19 & High \\
DexUMI~\citep{xu2025dexumi}         & 6  & 10 & 4  & Medium \\
RealDexUMI                          & 16 & 4  & 0  & Low \\
\bottomrule
\end{tabular}
\end{table}

\end{document}